\documentclass[runningheads]{llncs}
\usepackage[T1]{fontenc}
\usepackage{amsmath,amssymb,amsfonts}
\usepackage{algorithm}
\usepackage{algpseudocode}

\usepackage{graphicx,verbatim}
\usepackage{array}
\usepackage{utfsym}
\usepackage{booktabs}
\usepackage[caption=false,font=normalsize,labelfont=sf,textfont=sf]{subfig}
\usepackage{pifont}
\usepackage{makecell}
\usepackage{dsfont}
\usepackage{diagbox}
\usepackage{stfloats}
\usepackage{multirow}
\usepackage{verbatim}
\usepackage{graphicx}
\usepackage{textcomp}
\usepackage{orcidlink}
\usepackage{cite}
\usepackage{url}
\usepackage{color}
\begin{document}
\title{Prototype Instance-semantic Disentanglement with Low-rank Regularized Subspace Clustering for WSIs Explainable Recognition}
%

\author{Chentao Li, Pan Huang}  
\authorrunning{Chentao Li et al.}
\institute{Columbia University, Hong Kong Polytechnic University \\}
  
\maketitle              
\begin{abstract}
The tumor region plays a key role in pathological diagnosis. Tumor tissues are highly similar to precancerous lesions and non-tumor instances often greatly exceed tumor instances in whole-slide images (WSIs). These issues cause instance-semantic entanglement in multi-instance learning frameworks, degrading both model representation capability and interpretability. To address this, we propose an end-to-end prototype instance-semantic disentanglement framework with low-rank regularized subspace clustering, \textit{i.e.} PID-LRSC, in two aspects. First, we use secondary instance-subspace learning to construct low-rank regularized subspace clustering (LRSC), addressing instance entanglement caused by an excessive proportion of non-tumor instances. Second, we employ enhanced contrastive learning to design prototype instance‒semantic disentanglement (PID), resolving semantic entanglement caused by the high similarity between tumor and precancerous tissues. We conduct extensive experiments on multicentre pathology datasets, implying that PID-LRSC outperforms other state-of-the-art methods. Overall, PID-LRSC provides clearer instance semantics during decision-making and significantly enhances the reliability of auxiliary diagnostic outcomes. The code and material will be available \href{https://anonymous.4open.science/r/PID-LRSC-FB51}{\textit{here}}.

\end{abstract}
\section{Introduction}

 Tumor regions serve as the primary basis for pathological grading, as clinical decisions are largely determined within malignant tissue \cite{yan2021single}. However, in whole-slide images (WSIs), tumor areas account for only a very small proportion and they are highly similar to precancerous lesions in terms of morphological characteristics. This instance-semantic entanglement has intrinsically resulted in weak representation capability and limited interpretability of existing multi-instance learning (MIL) frameworks.

\begin{figure}
\centering
\includegraphics[width=4.8in]{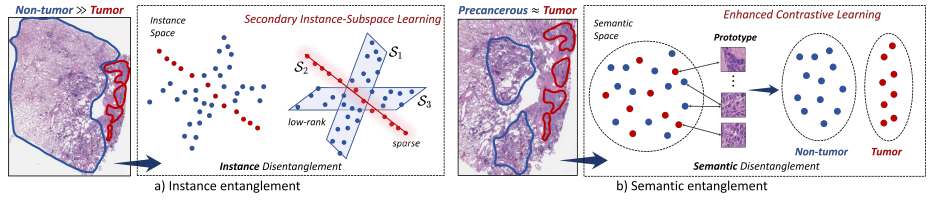}
\caption{The motivation of proposed PID-LRSC. \textbf{Left}: Instance entanglement caused by excessive proportion of non-tumor over tumor.  \textbf{Right}: Semantic entanglement caused by high similarity between precancerous lesions and tumor.}
\label{fig_1}
\end{figure}

Instance entanglement arises from the extreme imbalance within WSIs, where non-tumor regions overwhelmingly dominate but truly discriminative tumor instances are sparsely distributed \cite{kumar2020whole}. In Figure \ref{fig_1}, in the instance space, feature representations of non-tumor regions are highly redundant across different subspaces and thus noise accurate WSI representation. Without distinguishing and suppressing such irrelevant regions, the model fails to focus more on tumor-related subspaces and reduce instance-level subspace entanglement. For example, ACMIL \cite{zhang2024acmil} introduces multi-branch attention and stochastic top-k masking to capture diverse informative instances. DGR-MIL\cite{zhu2024dgrmil} employed a diverse global representation strategy to MIL framework  by modeling the diversity among instances through a set of global vectors. These models \cite{ilse2018abmil, lu2021clam, li2021dual, das2018multiple, lin2023ibmil, zhang2022dtfdmil, xiang2023ilramil, shao2021transmil,chikontwe2024frmil} ignore sparse tumor regions and are incapable of addressing instance entanglement. Meanwhile, biased MIL aggregation process will also overestimate redundant non-tumor instances and inhibit sparse tumor instances by assigning approximately equal weights, which weakens the model’s representational capacity and pathological grading performance.

Additionally, semantic entanglement stems from the high morphological similarity between precancerous lesions and malignant tumor cells, as illustrated in Figure \ref{fig_1}. In the semantic space, the ambiguity of tumor-related semantics significantly limits the interpretability of MIL models. Lack of external pathological criteria, the models cannot correctly identify the semantic category to which each instance belongs, failing to perform accurate posterior semantic attribution. Consequently, instance-level explanations become unreliable and may highlight irrelevant low-effect regions instead of truly discriminative tumor areas. For example, MiCo\cite{li2025mico} can enhance cross-regional intra-tissue correlations and strengthen inter-tissue semantic associations in WSIs with context-aware clustering. FuzzyMIL \cite{liu2025fuzzymil} proposed a deep fuzzy clustering framework based on learnable fuzzy C-means to analyze WSIs in a compact and efficient manner. These cluster-based MIL models \cite{wu2023improving, ju2023glcc, du2023neighbor, yang2023cluster, zheng2024partial, wang2023sac} have attempted to address instance entanglement by distinguishing subspaces, but they merely approximate instance pseudo labels and suffer from semantically entangled feature representation, degrading both interpretability and classification performance.

Incorporating external knowledge, prototype-driven learning can effectively alleviate poor model interpretability. PHIM-MIL \cite{xie2025phim} introduce prototypes for similar weight aggregation and multi-scale feature fusion. Gou et.al \cite{gou2025queryable} proposed a vision-language based queryable prototype MIL framework for incremental WSI classification. Moreover, HCF-MIL \cite{li2025hcfmil} is a prototype-driven MIL model with clustering to filter irrelevant non-tumor noise. These models either ignore the sparsity of tumor semantics for effective subspace clustering, or simply calculating prototype similarities, not exploring instance-semantic disentanglement.

For this purpose, we introduce PID-LRSC, an end-to-end instance-semantic disentangled learning framework for explainable WSIs recognition. First, we develop a low-rank regularized subspace clustering (LRSC) method for secondary instance-subspace exploration. Two coarse-to-fine low-rank constraints inhibit non-tumor features and enables sparse tumor features to form discriminative subspaces, alleviating entanglement caused by imbalanced instance distribution. Second, we implement prototype instance-semantic disentanglement (PID) with enhanced contrastive learning. By comparing with prototypes, the model tends to focus more on high-effect areas and learns semantically disentangled decision boundaries between tumor and precancerous lesions. Through end-to-end training, our PID-LRSC not only disentangles instances into domain-specific
semantics, but also improves the classification performance and interpretability.

\section{Methodology}
\subsection{Problem Statement}
Assume that differentiation states are hypothetically divided into tumor (TIs), non-tumor (NTIs) and background instances (BGIs) in a WSI, where tumor instances are in small proportion, yet are more correlated with pathological grading than non-tumor instances.

Instead of patch label fusion, We adopt instance aggregation based MIL framework that aims to learn a permutation-invariant scoring function $\mathcal{S}(\cdot)$ that maps the set of instances $\mathbf{X}=\{\boldsymbol{x}_1,\dots,\boldsymbol{x}_n\}$ to label $\mathbf{Y}$:
\begin{equation}
    \hat{\mathbf{Y}}=\mathcal{S}(\mathbf{X}) \quad\text{where}\quad \mathcal{S}(\mathbf{X}) =f\bigl(\mathcal{A}(\mathcal{F}(\mathbf{X})\bigr)
\end{equation}
where $\mathcal{A}(\cdot),f(\cdot)$ denote the aggregation function, prediction head with softmax normalization, respectively. $\mathcal{F}(\cdot)$ denotes the feature extractor.

\begin{figure*}[!t]
\centering
{\includegraphics[width=4.8in]{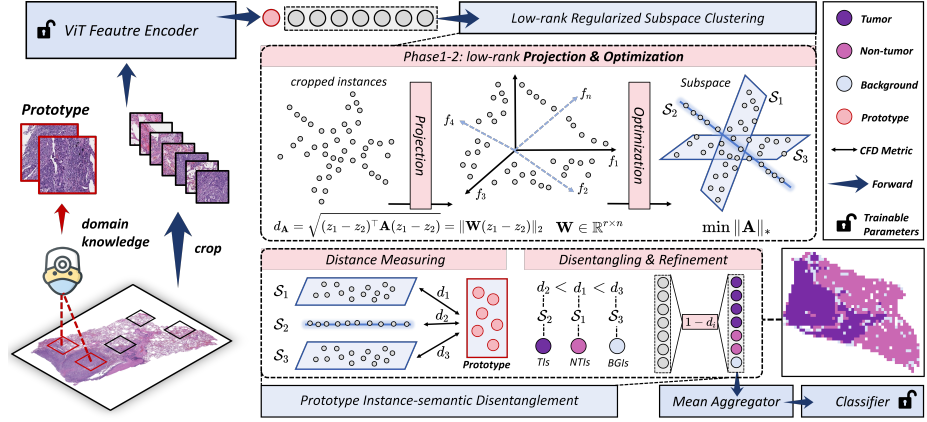}}
\hfil
\caption{Overview of \textbf{PID-LRSC}. Cropped instances extracted by encoder are clustered into three subspaces by \underline{\textbf{L}}ow \underline{\textbf{R}}ank \underline{\textbf{S}}subspace \underline{\textbf{C}}lustering method in two phases. They are then identified into tumor, non-tumor and background by \underline{\textbf{P}}rototype \underline{\textbf{I}}nstance-semantic \underline{\textbf{D}}isentanglement. End-to-end optimization is performed for disentangled representation learning with refined instance features.}
\label{fig:fig2}
\end{figure*}

\subsection{Two-phase Coarse-to-fine LRSC}
Sparse tumor regions are within large proportion of non-tumor regions in WSIs.
Inspired by deep clustering and metric learning, we proposed a two-phase coarse-to-fine adaptive clustering approach with low-rank constraint to obtain robust subspace results. The general form of metric learning is as followed:
\begin{equation}\label{eq2}
    \begin{aligned}
    d_{\mathbf{A}}(\boldsymbol{z}_1,\boldsymbol{z}_2)=\sqrt{(\boldsymbol{z}_1-\boldsymbol{z}_2)^T\mathbf{A}(\boldsymbol{z}_1-\boldsymbol{z}_2})
    \end{aligned}
\end{equation}
where $\mathbf{A}\in \mathbb{R}^{n\times n}$ is a trainable positive semi-definite matrix and $\boldsymbol{z}_1, \boldsymbol{z}_2 \in \mathbb{R}^{n\times 1}$ denote the instance feature representation vectors.

\noindent \textbf{Phase 1: Low-rank Projection}. Positive semi-definite matrix ensures the symmetry and non-negativity of the metric. It contributes to robust distance measuring in high dimensional space. Thus, we replace the $L_p$ norm with the adaptive metric $d(\cdot,\cdot)$ and ensure positive semi-definite\(\mathbf{A}\)  by parameterizing \(\mathbf{A} = \mathbf{W}^\top \mathbf{W}\), \(\mathbf{W}\in\mathbb{R}^{r\times n}\). Furthermore, we set \(r\ll n\) yielding a training-efficient metric. It also serves as a low-rank projection that searches for the best distribution-aware subspaces to obtain robust clustering results; see Fig. \ref{fig:fig2}.

\begin{equation}\label{eq3}
    \begin{aligned}
    d_{\mathbf{A}}(\boldsymbol{z}_1,\boldsymbol{z}_2)&=\sqrt{(\boldsymbol{z}_1-\boldsymbol{z}_2)^ \top\mathbf{W}^\top\mathbf{W}(\boldsymbol{z}_1-\boldsymbol{z}_2})=\|\mathbf{W}(\boldsymbol{z}_1-\boldsymbol{z}_2)\|_2
    \end{aligned}
\end{equation}
where $\text{rank}(\mathbf{A})\le\min(\text{rank}(\mathbf{\mathbf{W}}),\text{rank}(\mathbf{\mathbf{W^\top}}))\le r$. So $\mathbf{W}$ puts a hard low-rank constraint on $\mathbf{A}$ that directly and coarsely embeds features in a lower dimensional subspaces with less redundancy.

\noindent \textbf{Phase 2: Low-rank Optimization}. In Phase 1, we manually set a fixed $r$, but excessively small $r$ will restrict model's capability of representation. Since rank function is non-convex, we apply nuclear norm regularization as a soft constraint to guarantee rank minimization; see Fig. \ref{fig:fig2} and the following derivation.

\begin{equation}
    \begin{aligned}
        \|\mathbf{A}\|_*&=\sum_{i=1}^n\sigma_i(\mathbf{A})=\sum_{i=1}^n\sqrt{\lambda_i(\mathbf{A}^\top\mathbf{A})}=\sum_{i=1}^n\sqrt{\lambda_i(\mathbf{A}^2)}=\sum_{i=1}^n\sqrt{\Big (\lambda_i(\mathbf{A})\Big)^2}\\
    &=\sum_{i=1}^n|\lambda_i(\mathbf{A})|=\sum_{i=1}^n\lambda_i(\mathbf{A})=\text{Tr}(\mathbf{A})
    \end{aligned}
\end{equation}
where $\|\cdot\|_*$ denotes nuclear norm, and $\sigma_i(\cdot)$, $\lambda_i(\cdot)$ denote singular values, eigenvalues respectively. By minimizing this regularization term in the loss function, it fine-tunes the model to adaptively approximate the optimal rank associated with varying data distribution.

Overall, this coarse-to-fine LRSC incorporates the optimal \(\mathbf{A}^*=\mathbf{W}^{*\top}\mathbf{W}^*\) into KMeans, improving instance entanglement for TIs, NTIs, and BGIs.

\subsection{Latent Characteristic function discrepancy based PID}
After the LRSC clustering process, we obtain three subspaces, i.e. $\{\mathbf{Z}_1,\mathbf{Z}_2,\mathbf{Z}_3\}\leftarrow\mathbf{Z}=\mathcal{F}(\mathbf{X})$. We also denote prototype instance subspace $\mathbf{Z}^{\text{PIs}}$, which serves as the anchor to address instance-semantic entanglement. Conventional clustering-based MIL models perform mean-pooling or fixed weights for aggregation, which neglects sample diversities and instances distinctions. Consequently, a distribution-aware metric can cover more feature diversities for substantial samples.

\noindent \textbf{Distance Measuring}. Characteristic function discrepancy (CDF) is for robust assessment of distributional discrepancies in dataset distillation. It decouples both phase and amplitude information to ensure dataset realism and diversity \cite{wang2025ncfm}. Unlike maximum mean discrepancy (MMD), CFD enables flexible and latent kernel selection by solving a min-max problem:

\begin{equation}
    d_k=D(\mathbf{Z}_k,\mathbf{Z}^{\text{PIs}})=\mathbb{E}_{\boldsymbol{z}\sim \mathbf{Z}_k,\boldsymbol{z}'\sim \mathbf{Z}^{\text{PIs}}}\int_{\boldsymbol{t}} \sqrt{\text{Chf}(\boldsymbol{t})}dF_{\mathcal{T}}(\boldsymbol{t};\psi)
\end{equation}
\begin{equation}
    \begin{aligned}
    \notag
        \text{Chf}(\boldsymbol{t})= (|\Phi_{\boldsymbol{z}}(\boldsymbol{t})-\Phi_{\boldsymbol{z}'}(\boldsymbol{t})|)^2+\Big(2|\Phi_{\boldsymbol{z}}(\boldsymbol{t})||\Phi_{\boldsymbol{z}'}|\Big)\times\Big(1-\text{cos}(\boldsymbol{a}_{\boldsymbol{z}}(\boldsymbol{t})-\boldsymbol{a}_{\boldsymbol{z}'}(\boldsymbol{t}))\Big)
    \end{aligned}
\end{equation}
where $F_{\mathcal{T}}(\boldsymbol{t};\psi)$ is the sampling distribution on $\boldsymbol{t}$ with sampling function $\psi$ and $\boldsymbol{a}_{\boldsymbol{z}}(\boldsymbol{t})$, $\Phi_{\boldsymbol{z}}(\boldsymbol{t})$ are phase, amplitude function, respectively. Here $k=1,2,3$.

\noindent \textbf{Disentangling and Refinement}. We can disentangle the three clustered subspaces based on the CFD metric above and refine the WSI representation. The smallest distance $d_k$ means stronger effect of $\mathbf{Z}_k$ for SCC grading, i.e. tumor areas. Conversely, it is identified as background for the largest distance. Finally, we get the map with labels $\{\text{TIs},\text{NTIs},\text{BGIs}\}$ and employ the normalized distances to refine the corresponding instance features to get the ultimate representation $\mathbf{Z}_{\text{WSI}}$.
\begin{equation}\label{eq7}
    \mathbf{Z}_{\text{WSI}}=\Big(\mathbf{Z}^{\text{TIs}},\mathbf{Z}^{\text{NTIs}},\mathbf{Z}^{\text{BGIs}},\mathbf{Z}^{\text{PIs}}\Big)\cdot 
\begin{pmatrix}
1-d_{\text{min}} \\
1-d_{\text{med}} \\
1-d_{\text{max}} \\
1
\end{pmatrix}
\end{equation}
where $d_{\min},d_{\text{med}},d_{\max}$ are the minimum, median, maximum value of $d_k$.

\subsection{End-to-end Optimization}
End-to-end optimization can learn more domain-specfic feature representations. Specifically, we formulate the loss function as follows:
\begin{equation}
    \begin{cases}
         \mathcal{L}=\mathcal{L}_{ce}(f(\mathcal{A}(\mathbf{Z_{\text{WSI}}})))+\gamma_1 \cdot c_{\text{min}}-\gamma_2 \cdot c_{\text{max}}+\gamma_3 \cdot \text{Tr}(\mathbf{A})\\
        c_{\text{min}}=d_{\mathbf{A}}\Big(\text{mean}(\mathbf{Z}^{\text{TIs}}),\text{mean}(\mathbf{Z}^{\text{PIs}})\Big)\\
        c_{\text{max}}=d_{\mathbf{A}}\Big(\text{mean}(\mathbf{Z}^{\text{BGIs}}),\text{mean}(\mathbf{Z}^{\text{PIs}})\Big)\\
    \end{cases}
\end{equation}
where $\mathcal{L}_{ce}$ denotes the cross entropy. Since KMeans clustering is not normally differentiable, the cluster-separation regularizer $c_{\text{min}}$ and $c_{\text{max}}$ encourage tumor-instance cluster approach our prototype instances, optimizing the trainable matrix $\mathbf{A}$ by computing the gradients \(\frac{\partial \mathcal{L}}{\partial \mathbf{W}}\).

\section{Experimental Results}
\subsection{Datasets and Setup}
To validate PG-CIDL for diagnosis and sub-typing tasks, we conduct extensive experiments on two public datasets (DHMC-KIDNEY, DHMC-LUNG) and two private datasets (AMU-CSCC, AMU-LSCC). Details about datasets and implementation, please refer to Supplementary Material.

\begin{table*}[!t]
\small
\centering
\renewcommand\arraystretch{1.1}
\caption{Performance comparison of cancer diagnosis, and sub-typing on AMU-CSCC and AMU-LSCC with Swin-T pretrained features. DHMC-KIDNEY and DHMC-LUNG are for generalization.\label{tab:table1}}
\begin{tabular}{lcccccccc}
\Xhline{1pt}
\multirow{2}*{\normalsize Models} & \multicolumn{2}{c}{AMU-LSCC} & \multicolumn{2}{c}{AMU-CSCC} & \multicolumn{2}{c}{DHMC-KIDNEY} & \multicolumn{2}{c}{DHMC-LUNG}\\ \cmidrule(r){2-3} \cmidrule(r){4-5} \cmidrule(r){6-7} \cmidrule(r){8-9}
~ & \makebox[0.06\textwidth][c]{ACC}  & \makebox[0.06\textwidth][c]{AUC} & \makebox[0.06\textwidth][c]{ACC} & \makebox[0.06\textwidth][c]{AUC} & \makebox[0.06\textwidth][c]{ACC}  & \makebox[0.06\textwidth][c]{AUC} & \makebox[0.06\textwidth][c]{ACC}  & \makebox[0.06\textwidth][c]{AUC} \\ \hline
 ABMIL  \cite{ilse2018abmil}& 0.6739 & 0.7888 & 0.8302  & 0.9547 &0.7675&0.9421&0.7333&\underline{0.8123} \\
CLAM-SB \cite{lu2021clam}& 0.6957  & 0.8367 & 0.8302  & 0.9581 &0.7807&0.9339&0.7556&0.7783 \\
CLAM-MB \cite{lu2021clam}& 0.6812  & 0.8031 & 0.8774 & 0.9336 &0.7939&0.9372&0.7333&0.7684 \\
TransMIL \cite{shao2021transmil}& 0.8261  & 0.9021 & 0.8868 & 0.9521&0.8465&0.9603&0.7333&0.8084 \\
DTFD-MIL \cite{zhang2022dtfdmil}& 0.5435  & 0.7245 & 0.8113  & 0.9036 &0.7061&0.9159&0.7333&0.8084 \\
IBMIL-DS \cite{lin2023ibmil}& 0.6377  & 0.7672 & 0.7830  & 0.8881 &0.7939&0.9145&0.6889&0.7091 \\
ILRA-MIL \cite{xiang2023ilramil} & 0.6884  & 0.8048 & 0.8491  & 0.9123 &0.8333&0.8693&0.7778&0.8074  \\
S4MIL \cite{fillioux2023s4mil}& 0.7971 & 0.8845 & 0.8491  & 0.9605  &0.8158&0.9188&0.7111&0.7007\\
FRMIL \cite{chikontwe2024frmil} & 0.6377  & 0.8020 & 0.8113  & 0.8990 &0.7807&0.9174&0.7111&0.7442 \\
DGR-MIL \cite{zhu2024dgrmil}& 0.8188  & 0.9147 & 0.8868  & 0.9613 &\underline{0.8596}&\underline{0.9680}&\underline{0.8000}&0.7886  \\
ACMIL-MHA \cite{zhang2024acmil}& \underline{0.8551}  & \underline{0.9219} & \underline{0.9057}  & 0.9701  &0.8553&0.9560&0.6889&0.6849\\ 
RRT-MIL \cite{tang2024rrtmil}& 0.7681 & 0.8732 & 0.8585 & 0.9265  &0.8421&0.9378&0.7778&0.7521\\ 
MFC-MIL \cite{mfcmil2025} & 0.8261 & 0.9126 & 0.8962 & \underline{0.9723}  &0.8116&0.9456&0.7778&0.7767\\ \hline
\textbf{PID-LRSC (ours)} & \textbf{0.9130}  & \textbf{0.9590} & \textbf{0.9245}  & \textbf{0.9891} &\textbf{0.8991}&\textbf{0.9821}&\textbf{0.8222}&\textbf{0.8711}\\
\Xhline{1pt}
\end{tabular}
\end{table*}

\subsection{Comparison with SOTA Methods}
Table \ref{tab:table1} illustrates the sub-typing results of MIL approaches on two private datasets and two public datasets, which suggests that our proposed PID-LRSC has superior pathological grading performance. ACMIL-MHA achieves sub-optimal performance on AMU-LSCC, while PID-LRSC improves the ACC and AUC by 5.79\% and 3.71\%, respectively. For AMU-CSCC, the improvement is 1.88\% and 1.68\%. To further validate the capability of generalization, we compare those models on multicentre datasets. DGR-MIL achieves the second-best ACC and AUC on DMHC-KIDNEY, while our PID-LRSC improves them by 3.95\% and 1.41\%, respectively. DHMC-LUNG has seen similar trend with 2.22\% ACC improvement on AMU-CSCC and 5.88\% AUC improvement.

\begin{table}[t]
\small
\centering
\renewcommand\arraystretch{1.1}
\caption{Ablation experiments for different components of PID-LRSC on AMU-LSCC. \label{tab:table2}}
\begin{tabular}{cccccc}
\hline
\Xhline{0.8pt}
Models & Cluster & LRSC & PID & ACC & AUC\\ \hline
ACMIL-MHA & -- & -- & -- & 0.8551 & 0.9219\\
\multirow{4}*{PID-LRSC} & \multicolumn{1}{c}{\usym{1F5F4}} & \usym{1F5F4} & \usym{1F5F4} & \underline{0.8696} & 0.9567 \\
& \multicolumn{1}{c}{\usym{1F5F8}} & \usym{1F5F4} & \usym{1F5F4} & 0.8116 & 0.9455  \\
& \multicolumn{1}{c}{\usym{1F5F8}} & \usym{1F5F8} & \usym{1F5F4} & 0.8623 & \underline{0.9581}   \\
& \multicolumn{1}{c}{\usym{1F5F8}} & \usym{1F5F8} & \usym{1F5F8} & \textbf{0.9130} &\textbf{ 0.9590}  \\
\Xhline{0.8pt}
\hline
\end{tabular}
\end{table}

\begin{table}[!t]
\centering
\caption{Ablation experiments on different distribution metrics in PID.\label{tab:table3}}
\renewcommand\arraystretch{1.1}
\begin{tabular}{lcc}
\Xhline{0.8pt}
Metrics in PID (AMU-LSCC)&ACC & AUC\\ \hline
MMD & 0.8551 & 0.9441\\
\textbf{CFD (ours)} & \textbf{0.9130} &\textbf{0.9590} \\
\Xhline{0.8pt}
\end{tabular}
\end{table}

\begin{figure*}[!t]
\centering
{\includegraphics[width=4.8in]{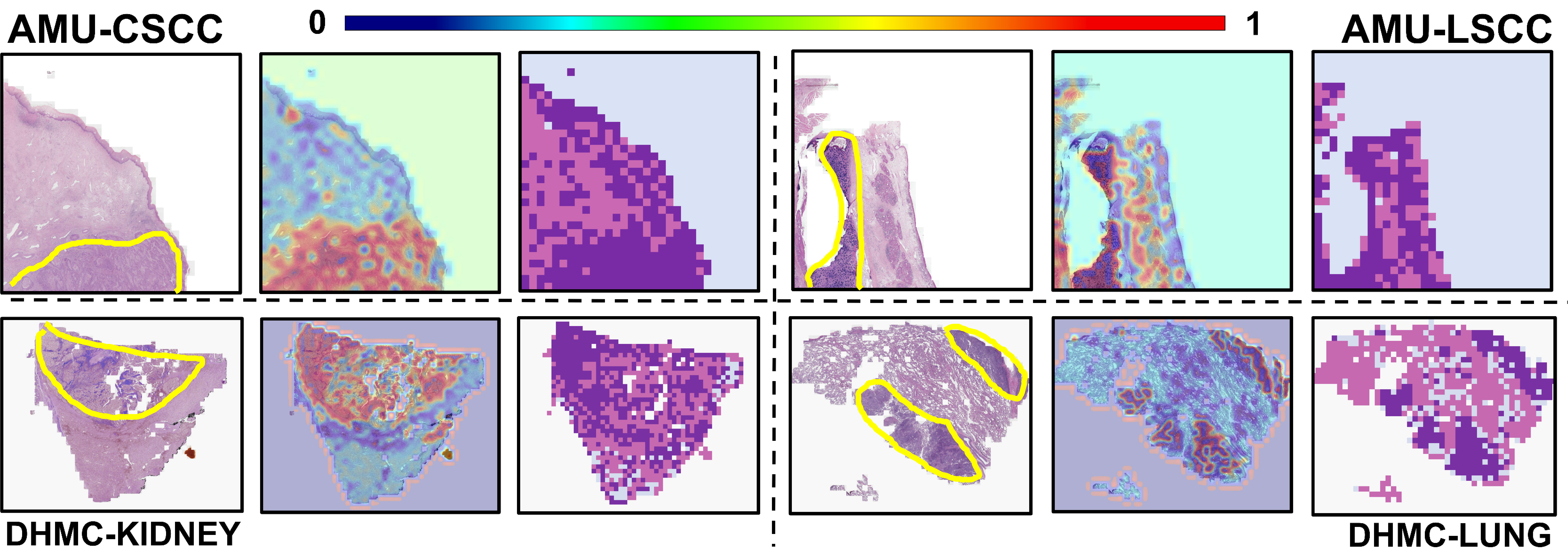}}
\hfil
\caption{Visualization results on multicentre datasets. The first column is original WSI with pathologist-annotated boundaries. The second is the attention contribution scores corresponding to its category. The last column is the disentangling results.}
\label{fig:fig4}
\end{figure*}

\subsection{Ablation Experiments}
In Table \ref{tab:table2}, without any component as Cluster, LRSC or PID, our baseline model achieves higher ACC and AUC compared to second-best ACMIL-MHA on AMU-LSCC. It indicates that end-to-end training with acceptable resources can learn better task-specific knowledge than two-stage framework. Moreover, directly incorporated L2-norm clustering in high dimensional space and fixed weights, the AUC performance only reaches 0.9455, 1.12\% drop compared to baseline. When LRSC is introduced to eliminate feature redundancy and robust clustering, the model performance improves significantly, with ACC increasing from 0.8116 to 0.8623 and AUC from 0.9455 to 0.9581. However, without the PID module, the clustering labels remain ambiguous. Therefore, by combining both the LRSC and PID modules, our model achieves the highest ACC 0.9130 and AUC 0.9590. Notice that we simply fix 3 clusters as for most WSI tasks, tumor, non-tumor tissue and background are most common differentiate stages. Larger cluster centroids are needed for more complex tasks \cite{song2024fasial}.

CFD enables flexible kernel for optimal distribution alignment, of which MMD is a special case. Therefore, we replace the MMD distance in PID module to show that the classification performance does degrade and CFD is a better distribution metric for WSI features with diversity and complexity.

\subsection{Visualization of WSI Interpretability and Representation}
We provide visualization results of our PID-LRSC on multicentre datasets in Figure \ref{fig:fig4}. We can tell that both disentangling results and attention heatmaps are highly aligned with pathologists' annotations.

\begin{figure*}[!t]
\centering
{\includegraphics[width=4.8in]{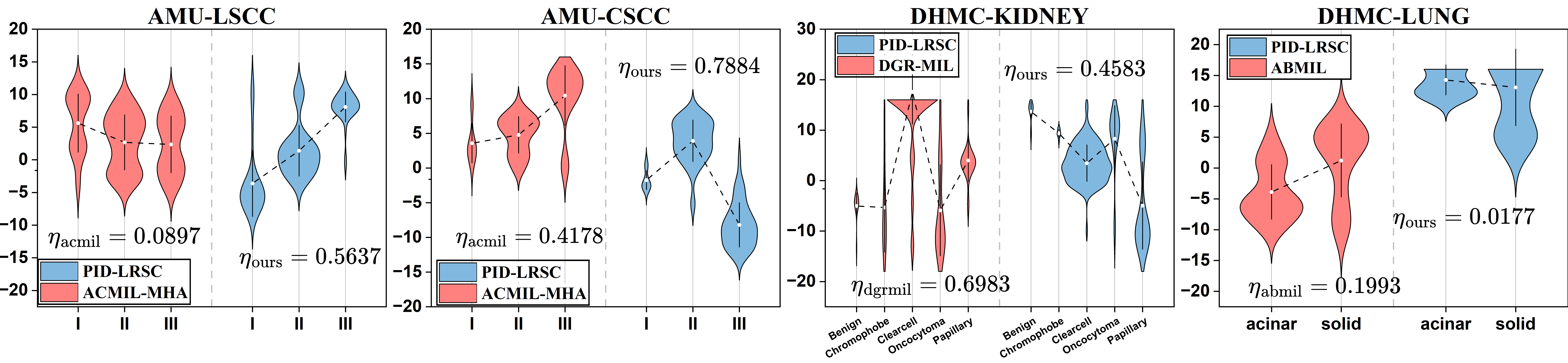}}
\hfil
\caption{ANOVA plots on four datasets where $\eta_*$ denote the value of effect size. Statistically, $\eta_*>0.4$ implies the model having large effects within categories.}
\label{fig:fig6}
\end{figure*}

An effective deep classification model is capable of capturing differences in feature distributions, which can be visualized through the analysis of variance (ANOVA) plots. Therefore, We employ UMAP to visualize the final layer feature representation of models in 1-D space. In ANOVA, a larger effect size $\eta^2$ indicates a stronger capacity of the model to capture and distinguish class-specific differences. Figure \ref{fig:fig6} demonstrates that PID-LRSC shows distinct distributions boundaries for SCC grading tasks, while ACMIL-MHA overlaps within different classes. Statistically, it has larger effect size 0.5637 and 0.7884 on both SCC datasets. Despite that DGR-MIL has larger $\eta$ value, our model still reaches a considerable level of effect. Therefore, we can conclude that our PID-LRSC has strong capability of feature representation.

\section{Conclusion}
Tumor regions play a pivotal role in pathological diagnosis, yet the high visual similarity between tumor and precancerous tissues, together with the overwhelming proportion of non-tumor instances in WSIs, leads to severe instance-semantic entanglement in MIL frameworks. Our proposed PID-LRSC, integrating low-rank regularized subspace clustering and  prototype instance-semantic disentanglement effectively disentangles semantic ambiguity between tumor and precancerous tissues. Extensive experiments on multicentre pathology datasets demonstrate that PID-LRSC consistently outperforms state-of-the-art methods, achieving superior classification performance and interpretability. Overall, PID-LRSC provides more explicit instance semantics and robust decision evidence, substantially enhancing the reliability and clinical value of WSI-based diagnosis.

\newpage

%
%
%
\bibliographystyle{splncs04}
\bibliography{my_ref}
%




\end{document}